\pdfoutput=1

\documentclass[11pt]{article}

\usepackage{emnlp2021}

\usepackage{times}
\usepackage{latexsym}

\usepackage[T1]{fontenc}

\usepackage[utf8]{inputenc}

\usepackage{graphicx}  
\usepackage{mathtools}  
\usepackage{amssymb}  
\usepackage{amsthm}  
\usepackage{multirow}  
\usepackage{subfigure}  
\usepackage{booktabs}  
\usepackage{algorithm}  
\usepackage{algcompatible}  
\usepackage{bm}  
\usepackage{xcolor}  
\newcommand{\red}[1]{\textcolor{red}{#1}}  
\newcommand{\itm}[1]{\textrm{\textit{#1}}}  
\usepackage{eqparbox}  
\newcommand{\down}{$\downarrow$}  

\usepackage{microtype}

\title{Transferable Persona-Grounded Dialogues via Grounded Minimal Edits}

\author{Chen Henry Wu\textsuperscript{\rm 1}, Yinhe Zheng\textsuperscript{\rm 2}, Xiaoxi Mao\textsuperscript{\rm 3}, Minlie Huang\textsuperscript{\rm 1} \\
\textsuperscript{\rm 1} Department of Computer Science and Technology, Institute for Artificial Intelligence, \\
State Key Lab of Intelligent Technology and Systems, Beijing National Research \\
Center for Information Science and Technology, Tsinghua University, Beijing, China \\
\textsuperscript{\rm 2} Samsung Research China - Beijing (SRC-B) \ 
\textsuperscript{\rm 3} Fuxi AI Lab, NetEase Inc., Hangzhou, China \\
{\normalsize\texttt{henrychenwu@cmu.edu, yh.zheng@samsung.com,}} \\
{\normalsize\texttt{maoxiaoxi@corp.netease.com, aihuang@tsinghua.edu.cn}} \\
}

\begin{document}
\maketitle
\begin{abstract}
Grounded dialogue models generate responses that are grounded on certain concepts. Limited by the distribution of grounded dialogue data, models trained on such data face the \textit{transferability} challenges in terms of the data distribution and the type of grounded concepts. To address the challenges, we propose the \textit{grounded minimal editing} framework, which minimally edits existing responses to be grounded on the given concept. Focusing on personas, we propose Grounded Minimal Editor (GME), which learns to edit by disentangling and recombining persona-related and persona-agnostic parts of the response. To evaluate persona-grounded minimal editing, we present the \textsc{PersonaMi-nEdit} dataset, and experimental results show that GME outperforms competitive baselines by a large margin. To evaluate the transferability, we experiment on the test set of \textsc{BlendedSkillTalk} and show that GME can edit dialogue models' responses to largely improve their persona consistency while preserving the use of knowledge and empathy.\footnote{Our codes and data are available at \url{https://github.com/thu-coai/grounded-minimal-edit}.}
\end{abstract}

\section{Introduction}
\label{sec:introduction}

Grounding dialogue agents on external information is important for building engaging conversational AI systems \cite{HuangZG20}. Along this track, various datasets and models have been proposed to ground dialogues on personas \cite{KielaWZDUS18}, knowledge \cite{DinanRSFAW19}, emotions \cite{Zhou2018EmotionalCM}, and images \cite{ShusterHBW20}.

\begin{figure}[t]
	\centering
	\includegraphics[width=0.9\linewidth]{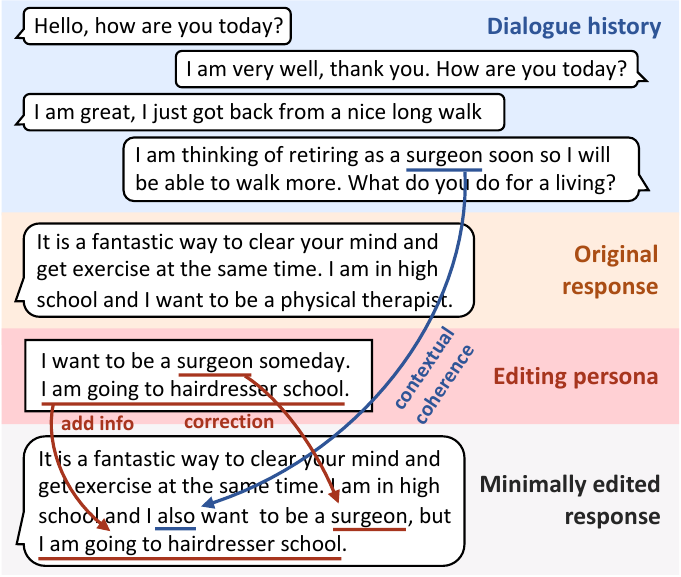}
	\caption{\label{fig:ucpt-example} Persona-grounded minimal editing. Edits are shown by arrows, accompanied by the explanations. }
\end{figure}

Generally, grounded dialogue modeling trains a dialogue model on a dataset $\mathcal{D}$ that consists of triples $(c, r, g)$, where $c$ is the dialogue history, $r$ is the response, and $g$ is the grounded concept. The model is generally optimized using maximum likelihood estimate (MLE), i.e., 
\begin{equation}
\label{eq:grounded-dialogue}
    \mathop{\arg\max}_{\theta}\mathbb{E}_{(c, r, g) \sim \mathcal{D}} \log P_{\theta}(r|c, g).
\end{equation}
Despite its effectiveness, this formulation faces two challenges regarding \textbf{\textit{transferability}}. On one hand, grounded dialogue datasets are usually collected under a guided setting, e.g., annotators are usually encouraged to embed persona \cite{KielaWZDUS18} or knowledge \cite{DinanRSFAW19} into responses, which leads to a \textbf{\textit{distributional gap}} between the conversations in a grounded dialogue dataset and natural conversations. As a result, models trained with Eq.~(\ref{eq:grounded-dialogue}) may generate unnatural responses and are vulnerable to the distributional shift of the dialogue history. On the other hand, at inference time, models trained with Eq.~(\ref{eq:grounded-dialogue}) cannot be grounded on unseen types of concept $g'$ other than $g$. An example for such \textbf{\textit{grounding gap}} is that a model trained on \textsc{PersonaChat} \cite{KielaWZDUS18} with Eq.~(\ref{eq:grounded-dialogue}) cannot be grounded on world knowledge. 

To address the above transferability challenges, we propose a \textit{grounded minimal editing} framework for grounded dialogue modeling. Instead of learning a \textit{grounded response generator} as is done in Eq.~(\ref{eq:grounded-dialogue}), we propose to learn a \textit{grounded minimal editor} that operates on existing responses. Specifically, suppose we have an original response $r^o$ that is coherent with the dialogue history $c$ but is not grounded on the concept $g$. Our goal is to \textit{minimally} edit $r^o$ such that it is grounded on the concept $g$ and coherent with the dialogue history $c$. Original responses can be generated by dialogue models trained on natural conversation data and grounded on other concepts $g'$, or even produced by humans; thus, they do not suffer from the \textit{distributional gap} and \textit{grounding gap}. Moreover, \textit{minimal} editing guarantees that the distribution of the edited responses is similar to that of the original responses, which do not suffer from the two gaps. Note that collecting paired responses before and after editing is resource-consuming; thus, our goal is to learn the editing \textit{without} paired data. 

In this paper, we explore \textit{persona-grounded minimal editing}, as demonstrated in Figure~\ref{fig:ucpt-example}. We propose Grounded Minimal Editor (GME), which is trained on persona-grounded dialogue data. Specifically, response templates are sampled by corrupting persona-related spans and sentences based on gradient-based attribution and word overlap. By denoising the templates, GME disentangles and recombines persona-related and persona-agnostic expressions. Since the personas of original responses are not observed at inference, we train a classifier for template generation at inference.

Two research questions are investigated in this paper: Q1) \textit{Is the proposed GME model effective for grounded minimal editing?} Q2) \textit{Does our framework address the transferability challenges (more specifically, the distributional gap and the grounding gap)?} For Q1, we build \textsc{PersonaMinEdit}, a new dataset derived from \textsc{PersonaChat} with multiple human references for the edited response. Automatic and human evaluations show that GME outperforms competitive baselines and has the most similar behavior to humans references. For Q2, we evaluate GME on the test set of \textsc{BlendedSkil-lTalk} \cite{SmithWSWB20}, whose data distribution and grounded concepts are different from \textsc{PersonaChat}, which requires GME to be transferable. We observe that GME improves the persona consistency of responses generated by pretrained Blender-90M models \cite{Roller20Blender}, while preserving the use of knowledge and empathy. Results also show that GME-edited responses largely outperforms TransferTransfo \cite{wolf19transfertransfo}, which is trained in the canonical way as in Eq.~(\ref{eq:grounded-dialogue}). Our contributions include: 

\begin{itemize}
    \item We propose a framework named \textit{grounded minimal editing} to address the transferability challenges of grounded dialogue modeling. 
    \item We propose Grounded Minimal Editor (GME) and present the \textsc{PersonaMinEdit} dataset to evaluate GME's effectiveness for persona-grounded minimal editing. 
    \item Experimental results show that GME largely outperforms strong baselines on the \textsc{PersonaMinEdit} dataset. GME is also transferable to edit other models' outputs and improve the persona consistency while preserving their use of knowledge and empathy.
\end{itemize}

\section{Related Work}
\label{sec:related-work}
Recent work leveraged grounded information in dialogue agents to chat engagingly, e.g., using knowledge \cite{Zhou2018Commonsense}, emotions \cite{Zhou2018EmotionalCM}, personas \cite{KielaWZDUS18}, and images \cite{ShusterHBW20}. For persona grounding \cite{LiGBSGD16,KielaWZDUS18}, transfer learning methods \cite{ZhangZWZL19,wolf19transfertransfo,GolovanovKNTTW19} and latent variable models \cite{SongZCWL19,ChanLYCHZY19} have shown promising results. Further, the persona consistency issue \cite{kim2020pragmatics,nie2020contradiction} and persona-augmented empathetic agents \cite{ZhongZWLM20} have also been explored. As discussed in Section~\ref{sec:introduction}, existing methods generally adopt the MLE objective in Eq.~(\ref{eq:grounded-dialogue}) and suffer from two transferability challenges, i.e., the distributional gap and the grounding gap, which are addressed by the proposed \textit{grounded minimal editing} framework. 

The idea of \textit{editing} existing responses has been explored, e.g., the deliberation network \cite{XiaTWLQYL17}, two-pass response generation \cite{SongWZLL20}, and retrieval-augmented dialogue modeling \cite{WestonDM18,ContractorKJP18,0006WHWL019,GuLZL19,CaiWBTLLS19}. This paper is essentially different from these works from two perspectives. 1) Regarding the formulation, we emphasize \textit{minimal} editing, while previous works do not. As analyzed in Section~\ref{sec:introduction}, \textit{minimal} editing is an important component to address the transferability challenges; 2) Regarding the training algorithm, previous works derive templates from self-generated or retrieved texts, while our model derives templates from the observed responses.

Our work is also related to controlled text editing without parallel data, e.g., unsupervised text style transfer \cite{shen2017style,LiJHL18,RaoT18,LampleSSDRB19}, semi-supervised contextual text style transfer \cite{Cheng20Contextual}, syntax-controlled paraphrasing \cite{bao2019syntax}, contrastive model explanation \cite{Ross20Mice}, counterfactual story generation \cite{QinBHBCC19,QinSWBHBBC20}, and sentence-level editing for empathetic dialogues \cite{sharma2021facilitating}. Some of these studies also utilize masked templates \cite{LiJHL18,WuZZHH19,SudhakarUM19,MalmiSR20,Ross20Mice}. However, these previous works only focus on categorical conditions in a small label space, while the personas in our study are embedded in much larger spaces. In the large persona space, the persona sentences at test time are never seen during training. Further, when generating masked templates, the personas of the original responses are unobserved in our study. 


\section{Formulation}
\label{sec:formulation}

\begin{figure}[t]
	\centering
	\includegraphics[width=0.8\linewidth]{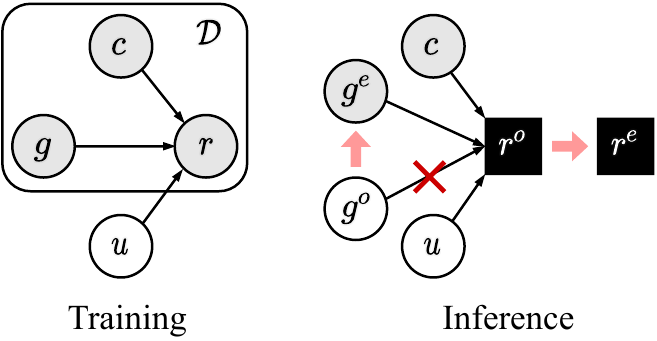}
	\caption{\label{fig:formulation} Graphical formulation of grounded minimal editing. Observed variables are shown in grey, and unobserved variables are shown in white ($c$: dialogue history, $g$: grounding, $r$: response, $\mathcal{D}$: training data, $u$: unobserved variables). At inference, the editing $r^{o} \rightarrow r^{e}$ is based on $g^{o} \rightarrow g^{e}$, while the unobserved variables $u$ remain unchanged. Note that $g^{o}$ is also not observed.}
\end{figure}

We provide a formulation of the proposed framework. Grounded dialogue modeling uses a dataset $\mathcal{D}$ that consists of triples $(c, r, g)$, where $c$, $r$, and $g$ are the dialogue history, the response, and the grounded concept, which are shown in grey in the left part of Figure~\ref{fig:formulation}. To formulate the term ``\textit{minimal}'', we need to add \textit{unobserved variables} into the graphical model, denoted as $u$ in Figure~\ref{fig:formulation}, which cover \textit{all} unobserved variables. The graph states that $r = f(c, g, u)$. As shown in the right part of Figure~\ref{fig:formulation}, we observe $(c, r^{o}, g^{e})$ at inference time, where $r^{o}$ and $g^{e}$ stand for the original response and the grounded concept for editing. The graph states that the original response $r^{o} = f(c, g^{o}, u)$, where $g^{o}$ represents the concept the original response is grounded on, and that both $g^{o}$ and $u$ are unobserved. The edited response is defined as $r^{e} = f(c, g^{e}, u)$, which replaces $g^{o}$ as $g^{e}$, and keeps $c$ and $u$ intact. Our formulation follows the idea of counterfactual reasoning \cite{PetersJanzingSchoelkopf17}, and it guarantees that 1) the content irrelevant to the grounded concept is preserved, and that 2) the edited response is coherent with the dialogue history. Since it is costly to collect paired $(r^{o}, r^{e})$ for training, the grounded minimal editor should be trained on the grounded dialogue data $(c, r, g) \sim \mathcal{D}$ as in Eq.~(\ref{eq:grounded-dialogue}). 

As the first attempt toward the proposed framework, we focus on \textit{persona-grounded} minimal editing in the experiments. Thus, in the remaining part of this paper, we set the grounded concept $g$, $g^{o}$, $g^{e}$ as the persona $p$, $p^{o}$, $p^{e}$. 

\section{Our Approach}
\label{sec:method}

\subsection{Overview}
We propose Grounded Minimal Editor (GME), a pipeline model for grounded minimal editing. At inference, GME first creates a \textit{response template} $t$ by masking persona-related spans in the original response $r^{o}$ and then \textit{recombines} the template $t$, the persona $p^{e}$, and the dialogue history $c$ into an edited response $r^{e}$. We design the template to approximate the unobserved variables $u$ in Section~\ref{sec:formulation}, which distinguishes GME from previous retrieval-based dialogue models. With some abuse of notation, we use $t$ to denote the template for both training and inference. During training, two modules are learned: 1) a generator used for the recombination described above and 2) a mask classifier that helps create the response template at inference. Note that GME can also be applied to other ground concepts besides personas. The full process is presented in Algorithm~\ref{alg:training-inference}.

\begin{figure*}[t]
	\centering
	\includegraphics[width=0.95\linewidth]{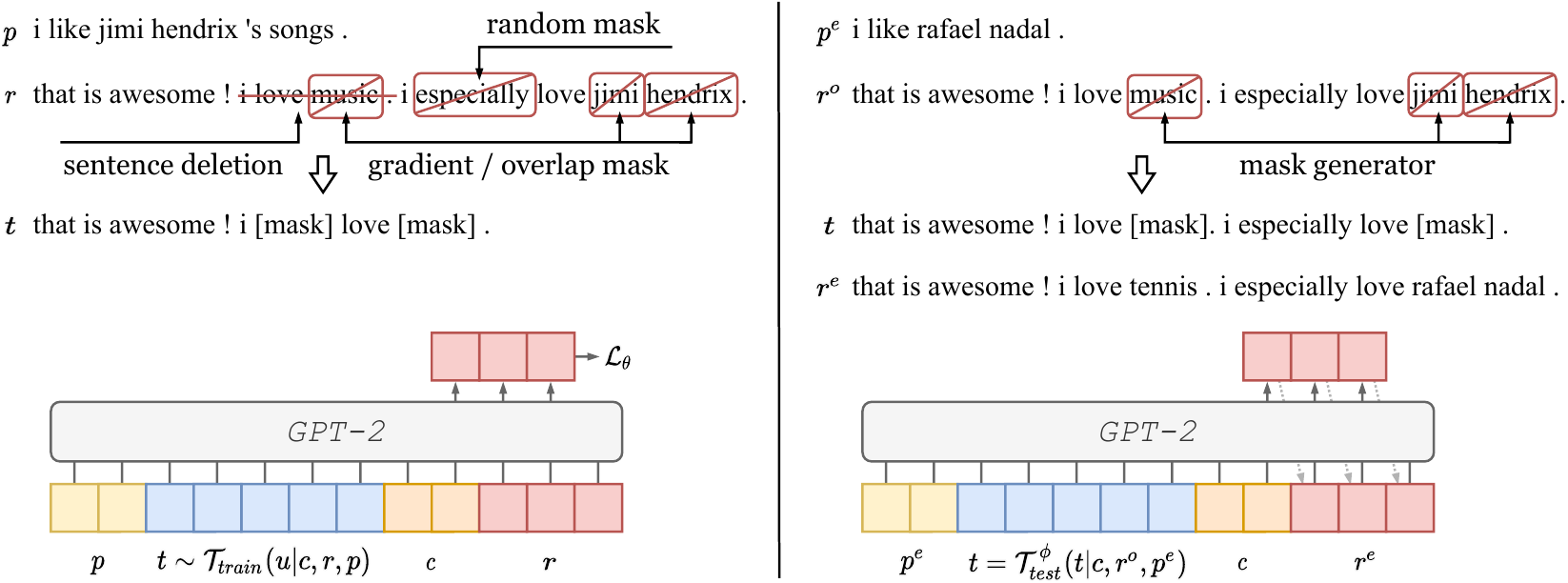}
	\caption{\label{fig:model-specification} Left: training example and input format. Right: inference example and input format. For readability, the same response sample is used. The context, position embeddings, and token type embeddings are omitted here.}
\end{figure*}

\subsection{Recombination Module}
The recombination module learns to recombine the response template, the persona, and the dialogue history as the edited response. During training, we create templates from the training responses, as detailed below. 

\paragraph{Span mask} \ The span mask serves as the placeholder of persona-related spans. For each response-persona pair, we define three sets of tokens: \textsc{Gradient}, \textsc{Overlap}, and \textsc{Stopwords}. \textsc{Gradient} contains persona-related tokens that are determined using \textit{gradient-based attribution} \cite{SimonyanVZ13}. We pretrain a response-to-persona model and compute the $L_{2}$ norm of the gradient of the persona's cross-entropy loss w.r.t. each response token's embeddings. A token is placed into the \textsc{Gradient} set if the $L_{2}$ norm is greater than $\delta=3$. \textsc{Overlap} contains response tokens whose lemma overlaps with in the lemmas of the persona tokens, which are likely to be related to the persona. \textsc{Stopwords} contains stopwords and punctuation marks specified by NLTK \cite{Bird06}. We mask a token if it is in \textsc{Gradient} or \textsc{Overlap} but not in \textsc{Stopwords}. We call sentences with masks after this step as \textit{persona-related sentences}. For each persona-related sentence, we further mask 15\% of its tokens to improve the robustness. Since the number of tokens varies at the same syntactic position, we merge consecutive masks so that all masks are at the span level.  

\begin{algorithm}[t]
\small
   \caption{\ Training and inference of GME}
   \label{alg:training-inference}
\begin{algorithmic}[1]
\STATEx // \textbf{Training}
\REPEAT \ sample $(c, r, p) \sim \mathcal{D}$
    \STATE Sample $t \sim \mathcal{T}_{\itm{train}}(t|r, p)$
    \STATE Optimize $\mathcal{L}_{\theta}$ in Eq.~(\ref{eq:denoising-objective}) and $\mathcal{L}_{\phi}$ in Eq.~(\ref{eq:classification-loss})
\UNTIL convergence
\STATEx // \textbf{Inference}
\STATEx \textbf{Input}: $(c, r^{o}, p^{e})$
\STATE Infer $t = \mathcal{T}^{\phi}_{\itm{test}}(t|c, r^{o}, p^{e})$
\STATE Edited response $r^{e} \sim P_{\theta}(r^{e}|c, t, p^{e})$
\end{algorithmic}
\end{algorithm}

\paragraph{Sentence deletion} The above span mask is effective for correcting persona \textit{contradictions} in the original response. However, span mask cannot handle the situation where we want to add new persona information into the response (examples are given in Figure~\ref{fig:ucpt-example} and 
Appendix~E). 
To model this pattern, we randomly delete persona-related sentences. Suppose we have $l$ persona-related sentences in the response, the number to keep $0 \leq n \leq l-1$ follows $P(n) \propto \exp(-n / \tau)$, where $\tau$ is a hyperparameter. By infilling persona-related sentences, the model learns to merge persona into the response. 

\paragraph{}

An example of the training template is shown in Figure~\ref{fig:model-specification}. During training, the recombinition modules $P_{\theta}$ is optimized by
\begin{equation}
\label{eq:denoising-objective}
\begin{split}
    \mathcal{L}_{\theta} = - \mathbb{E}_{(c, r, p) \sim \mathcal{D}}&[\mathbb{E}_{t \sim \mathcal{T}(t|r, p)} \log P_{\theta}(r|c, t, p)]. \\
\end{split}
\end{equation}
where $\mathcal{T}(t|r, p)$ denotes the distribution of the template as detailed above. As shown in Figure~\ref{fig:model-specification}, we use GPT-2 as the backbone to parameterize $P_{\theta}$, which is tackled as a language modeling task by concatenating input texts. We apply label smoothing ($\epsilon = 0.1$), and we use greedy decoding at inference. Token type embeddings are used to distinguish each type of text and each speaker. 

\subsection{Mask Generator}
\label{subsec:mask-generator} 

Since the persona of the original response before editing, i.e., $p^{o}$, is unobserved at inference, we train a mask generator $P_{\phi}$ to predict if a token $r_{i}$ should be masked. The objective for the mask generator is
\begin{equation}
\label{eq:classification-loss}
    \mathcal{L}_{\phi} = - \mathbb{E}_{(c, r, p) \sim \mathcal{D}}\sum_{i = 1}^{|r|}\frac{1}{f_{i}}\log P_{\phi}(m_{i}|c, r)
\end{equation}
where $m_{i}=1$ if $r_{i}$ is in \textsc{Gradient} or \textsc{Overlap} but not in \textsc{Stopwords}, and $m_{i}=0$ otherwise. $f_{i}$ is the corpus-level frequency of $m_{i}$, which is used to balance the number of positive samples and negative samples. At inference, we mask a word if 1) $P_{\phi}$ labels it as \textit{masked} with a confidence greater than $\epsilon$ ($\epsilon=0.5$ in the main experiment, $\epsilon=0.75$ in the transferability experiment) and meanwhile 2) it does not appear in the persona $p^e$ or the dialogue history $c$. We merge consecutive masks to get span masks. This process is denoted as $\mathcal{T}^{\phi}_{\itm{test}}(t|c, r^{o}, p^{e})$ in Algorithm~\ref{alg:training-inference}. 

\section{Evaluation Data: \textsc{PersonaMinEdit}}

\subsection{Data Collection}
\label{subsec:benchmark}
We present a new dataset \textsc{PersonaMinEdit} to evaluate persona-grounded minimal editing. Validation and test data are collected in two steps:

\noindent\textbf{Editing persona selection \ \ } We first construct inference samples $(c, r^{o}, p^{e})$, where the dialogue history $c$ and original response $r^{o}$ are from \textsc{PersonaChat}, and we select the editing persona $p^{e}$ based on two criteria: 1) \textit{editing difficulty} and 2) \textit{conversation consistency}. We bias our data to the hard cases that require correction of \textit{persona contradictions}. 
Specifically, we use the heuristics provided by \citet{WelleckWSC19} to select personas that are contradictory to the original response. To ensure conversation consistency, we filter out personas that are contradictory to the speaker's responses in the dialogue history. Finally, we also ensure that the persona sentences within each persona are not contradictory to each other. 

\noindent\textbf{Response editing \ \ } For each constructed triple $(c, r^{o}, p^{e})$, we collect references for the edited responses $r^{e}$ on Amazon Mechanical Turk. Specifically, $r^{e}$ should satisfy three requirements: 1) consistency with the editing persona $p^{e}$, 2) minimal editing, and 3) coherence with the dialogue history $c$. We reject annotations that do not add words to the original response. \textit{Three} human references are collected for each triple, and duplicate references are re-annotated. The inter-annotator BLEU (i.e., the BLEU of each reference given the other two references) is 73.8 on the validation set and 71.4 on the test set. The annotation instructions we used are detailed in 
Appendix~A.

Training data in \textsc{PersonaMinEdit} is derived from the training data of \textsc{PersonaChat}, and personas are aligned with responses following \citet{WelleckWSC19}. We also \textit{remove training samples whose persona appears in the editing personas in the validation and test data} to ensure that the \textit{\textbf{persona does not leak}} from training to testing. 


\begin{table}[t]
\centering
\small
\begin{tabular}{@{}lccccc@{}}
\toprule
~  	            & \textit{add}   & \textit{rm}    & $\Delta L$  & $d(r^{e}, r^{o})$ & $d(r^{e}, p^{e})$ \\
\midrule
Valid           & 5.1  & 2.7      & $+$ 3.2 & 7.0 & 11.0 \\
Test            & 4.9  & 2.6      & $+$ 2.9 & 6.7 & 10.8 \\
\bottomrule
\end{tabular}
\caption{\label{tab:data-analysis} Data analysis. Notations follow Section~\ref{subsec:data-analysis}. }
\end{table}

\subsection{Data Statistics}
\label{subsec:data-analysis}

After removing training samples whose persona appears in the editing personas in the validation and test splits, our training data has 119,078 samples. The validation split has 1,384 samples (1,266 with one sentence in the editing persona, 118 with two). The test split also has 1,384 samples (1,269 with one sentence in the editing persona, 115 with two). 

We study the \textit{behavior} of human references to understand the human intuition of minimal editing. In Table~\ref{tab:data-analysis}, we report the number of words added (\textit{add}) and removed (\textit{rm}), and the length difference ($\Delta L$) between the edited and original responses. We also report the minimum edit distance (MED) between the edited and original responses ($d(r^{e}, r^{o})$), and that between the edited response and the editing persona ($d(r^{e}, p^{e})$). We observe that the edited responses are generally local modifications of the original responses. On average, the edited responses are longer than the original ones, which can be explained by the observation that human sometimes add persona information into the response when no persona contradiction exists. 

\section{Experiment on \textsc{PersonaMinEdit}}
\label{sec:experiment-grounded-editing}

We use \textsc{PersonaMinEdit} to evaluate persona-grounded minimal editing (Q1 in Section~\ref{sec:introduction}). 

\subsection{Baselines}
\label{subsec:grounded-editing-baselines}

We modify state-of-the-art models for unsupervised text style transfer and counterfactual story generation as the baselines for grounded minimal editing. 

\noindent\textbf{No edit \ \ \ } This baseline does not make any edits to the original response. 

\noindent\textbf{UNMT \ \ \ } \citet{LampleSSDRB19,HeWNB20} adopted the unsupervised neural machine translation (UNMT) model \cite{LampleOCDR18} for unsupervised text style transfer. For our task, we replace the \textit{style condition} with \textit{persona condition}, and use a word dropout rate $p_{\itm{wd}} \in \{0.1, 0.5\}$. 

\noindent\textbf{CycleGAN \ \ \ } \citet{LuoLZYCSS19,DaiLQH19} adopted CycleGAN \cite{ZhuPIE17} for unsupervised text style transfer. For our task, we replace the \textit{style classifier} with a \textit{response-to-persona model}. We use Gumbel-softmax straight through gradient estimator \cite{JangGP17} for optimization.



\noindent\textbf{DeLorean-FT \ \ } DeLorean \cite{QinSWBHBBC20} iteratively modifying GPT-2's logits via gradients from a content preserving loss. For our task, we replace GPT-2 with TransferTransfo \cite{wolf19transfertransfo} and set the mixture rate $\gamma_{\itm{mix}} \in \{0.75, 0.80, 0.85\}$, where larger (smaller) $\gamma_{\itm{mix}}$ is biased towards persona consistency (minimality of editing). 

We observe that CycleGAN is sensitive to hyperparameters and unstable to train, probably due to the biased gradient estimation given the large persona space. Thus, we do not include other methods that require gradient backpropagation from classifiers \cite{ZhouCLXSGW20,madaan2020generate}. 

\subsection{Automatic Evaluation}
\label{subsec:grounded-editing-automatic-evaluation}
For automatic evaluation, we run each experiment with five random seeds. More details are presented in 
Appendix~C.

\noindent\textbf{BLEU\ \ \ } We compute BLEU-4 score \cite{PapineniRWZ02} based on the collected multiple human references, using the Moses script \textit{multi-bleu.perl}. From Table~\ref{tab:automatic-results} and Table~\ref{tab:behavioral-analysis}, we observe that higher BLEU indicates the less editing. 

\noindent\textbf{P-Score\ \ } We define P-Score to evaluate the persona consistency. Specifically, we finetune a BERT model on the DNLI dataset \cite{WelleckWSC19} to predict the relation $C(r, p_{j})$ (\textit{entailment}, \textit{neutral}, or \textit{contradiction}) of a response $r$ and a persona sentence $p_{j}$.\footnote{We use the classifier provided by \cite{MadottoLWF19}, which has 92.57\% accuracy on the DNLI verified test set.} We then map \textit{entailment}, \textit{neutral}, and \textit{contradiction} to $+0.5$, $0$, and $-0.5$ and define the P-Score of a sample as
\begin{equation}
\label{eq:p-score}
    \textrm{P-Score} = \sum_{j} \itm{map}[C(r^{e}, p^{e}_{j})]
\end{equation}
where $r^{e}$ is the edited response and $p^{e}_j$ is a persona sentence in $p^{e}$. We finally report the P-Score averaged over all samples.

\noindent\textbf{Average\ \ } We observe that BLEU and P-Score show a trade-off between minimal editing and persona consistency. We report their arithmetic mean as the overall performance since BLEU and P-Score have similar scales and variances. 

\begin{figure}[t]
	\centering
	\includegraphics[width=0.95\linewidth]{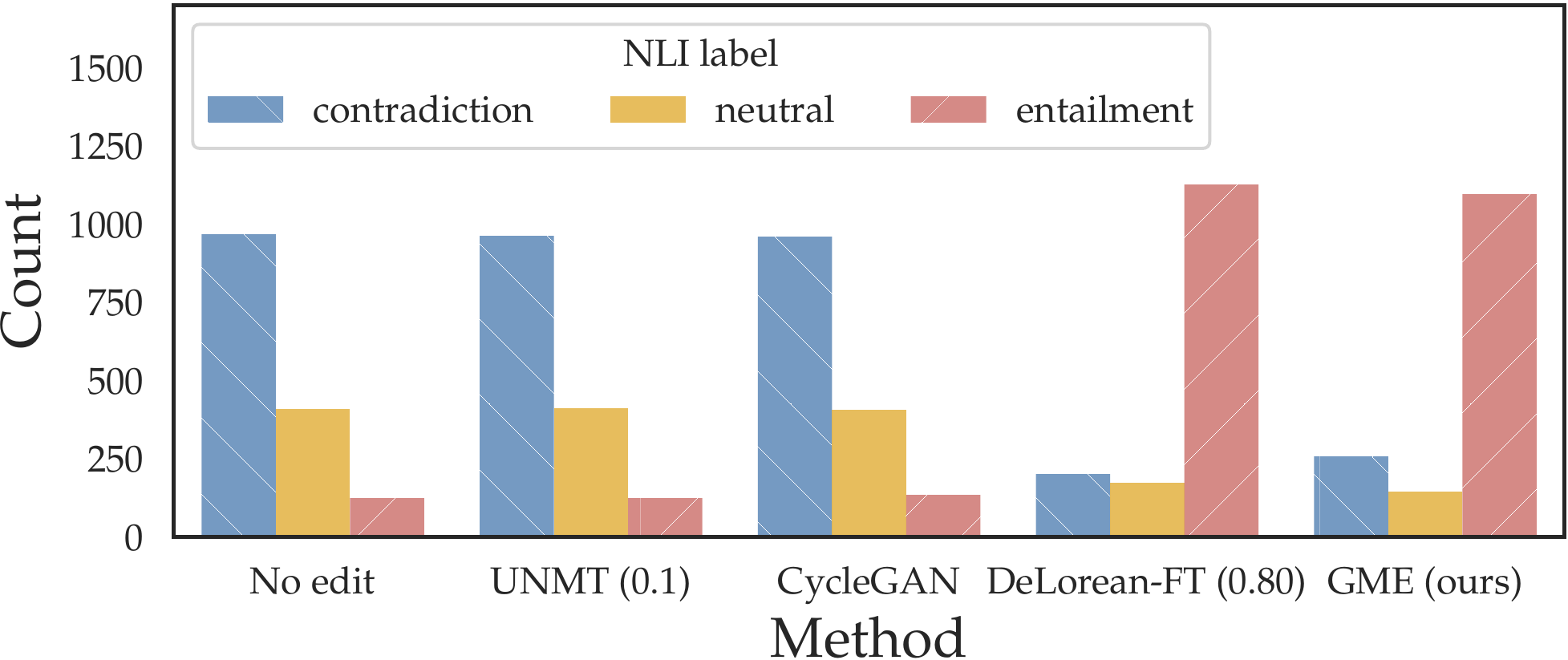}
	\caption{\label{fig:p-score} Distribution of classes in P-Score: contradiction (blue backslash), neutral (yellow), and entailment (red slash). Hyperparameters are in parenthesis.}
\end{figure}

Table~\ref{tab:automatic-results} shows that CycleGAN and UNMT have high BLEU but negative P-Scores. Figure~\ref{fig:p-score} shows that most of their outputs are contradictory to the editing personas, indicating that their edits are not focused on persona-related expressions. These results show that methods designed for binary style labels are not effective for persona-grounded minimal editing, where the persona space is much larger than the label space.  Larger $\gamma_{\itm{mix}}$ for DeLorean-FT lead to lower BLEU and higher P-Score, showing that larger (smaller) $\gamma_{\itm{mix}}$ is biased towards persona consistency (minimality of editing). However, results show that the overall performance cannot be improved by hyperparameter tuning. 

GME achieves a $31.9\%$ relative improvement on the Average score over the best performing baseline (from $34.2$ to $45.1$). Figure~\ref{fig:p-score} shows that most of GME's outputs entail the given personas. Table~\ref{tab:ablation-studies} shows the results for 1) removing \textit{dialogue histories} from the data and 2) removing \textit{sentence deletion} from GME. We observe that the dialogue history only has a slight contribution, showing that the response template contains an adequate amount of information of the original response. Sentence deletion contributes largely to the performance, especially for the persona consistency.

\begin{table}[t]
\centering
\small
\begin{tabular}{@{}lccc@{}}
\toprule
~  			                            & BLEU          & P-Score           & Average \\
\midrule
No edit                                 & 76.4 (0.0)    & $-$ 30.5 (0.0)    & 23.0 (0.0) \\ 
UNMT \\                 
\quad -- $p_{wd}=0.1$                   & 74.2 (0.2)    & $-$ 30.2 (0.3)    & 22.0 (0.2) \\ 
\quad -- $p_{wd}=0.5$                   & 69.0 (0.2)    & $-$ 27.9 (0.7)    & 20.6 (0.4) \\ 
CycleGAN                                & 74.4 (0.8)    & $-$ 28.3 (1.6)    & 23.0 (0.7) \\ 
DeLorean-FT \ \ \\ 
\quad -- $\gamma_{\itm{mix}}=0.75$      & 39.8 (2.2)    & $+$ 26.4 (2.8)    & 33.1 (0.6) \\ 
\quad -- $\gamma_{\itm{mix}}=0.80$      & 34.5 (0.7)    & $+$ 32.6 (1.6)    & 33.5 (0.6) \\ 
\quad -- $\gamma_{\itm{mix}}=0.85$      & 32.0 (0.8)    & $+$ 36.5 (1.0)    & 34.2 (0.7) \\ 
\midrule
GME (ours)                              & 60.3 (1.8)    & $+$ 29.9 (2.2)    & \textbf{45.1} (0.5) \\ 
\bottomrule
\end{tabular}
\caption{\label{tab:automatic-results} Automatic evaluation. We report the average of 5 random seeds, and standard deviations are shown in parenthesis. Details of P-Score are in Figure~\ref{fig:p-score}.}
\end{table}

\begin{table}[t]
\centering
\small
\begin{tabular}{@{}l@{}c@{}c@{}c@{}}
\toprule
Baseline                        & \textit{prefer baseline}  & \ \ \ \ \ \ \textit{none} \ \ \ \ \ \ & \textit{prefer GME} \\
\midrule
UNMT ($0.1$)                    & 0.0 \%                    & 59.3 \%       & \textbf{40.7} \% \\
CycleGAN                        & 0.7 \%                    & 59.1 \%       & \textbf{40.2} \% \\
DeLorean-FT ($0.85$)            & 8.4 \%                    & 52.7 \%       & \textbf{38.9} \% \\
\bottomrule
\end{tabular}
\caption{\label{tab:human-results} Human evaluation. Free-marginal $\kappa$ for each row is 0.66, 0.66, and 0.51 (\textit{substantial, substantial, and moderate agreement}).}
\end{table}

\begin{table}[t]
\centering
\small
\begin{tabular}{@{}lccc@{}}
\toprule
~  			                  & BLEU        & P-Score           & Average \\
\midrule
Full                          & 60.3 (1.8)  & $+$ 29.9 (2.2)    & 45.1 (0.5) \\
\midrule
w/o history                   & 60.2 (1.0)  & $+$ 29.3 (2.0)    & 44.8 (0.8) \\ 
w/o sent. del.                & 64.2 (0.3)  & $+$ 11.0 (0.3)    & 37.6 (0.2) \\ 
\bottomrule
\end{tabular}
\caption{\label{tab:ablation-studies} Ablation studies. Notations follow Table~\ref{tab:automatic-results}. }
\end{table}

\begin{table}[t]
\centering
\small
\begin{tabular}{@{}l@{}c@{}c@{}c@{}c@{}c@{}c@{}c@{}}
\toprule
& \ \ \textit{add} \ \ & \ \ \ \textit{rm} \ \ \ & \ \ \ $\Delta L$ \ \ & \ \ $d(r^{e}, r^{o})$ \ \ & \ $d(r^{e}, p^{e})$ \\
\midrule
No edit                             & 0.0  & 0.0    & \ 0.0   & 0.0  & \bf 12.0 \\
UNMT ($0.1$)                        & 0.2  & 0.1    & $+$ 0.1 & 0.2  & 12.1 \\
UNMT ($0.3$)                        & 0.6  & 0.4    & $+$ 0.3 & 1.0  & 12.2 \\
CycleGAN                            & 0.1  & 0.1    & $+$ 0.1 & 0.2  & 12.1 \\
DeLorean-FT \ \ \ \ \ \ \ \  \\ 
\quad -- $\gamma_{\itm{mix}}=0.75$  & \bf 5.7  & 7.9    & $-$ 1.7 & 9.8  & 7.5 \\
\quad -- $\gamma_{\itm{mix}}=0.80$  & 6.3  & 8.7    & $-$ 1.8 & 10.9 & 7.2 \\
\quad -- $\gamma_{\itm{mix}}=0.85$  & 6.9  & 9.2    & $-$ 1.7 & 11.7 & 7.0 \\
\midrule
GME (ours)                          & \bf 4.0  & \bf 2.5    & $+$ \bf 3.6 & \bf 6.7  & 13.0 \\
\midrule
Human references                    & 4.9  & 2.6    & $+$ 2.9 & 6.7  & 10.8 \\
\bottomrule
\end{tabular}
\caption{\label{tab:behavioral-analysis} Behavioral statistics. Results that are \textbf{the closest to human references} are shown in bold.}
\end{table}

\subsection{Human Evaluation} 
We randomly sample 150 test samples for human evaluation. Given two edited responses \textit{A} and \textit{B}, three annotators are hired to vote \textbf{\textit{prefer A}}, \textbf{\textit{none}}, or \textbf{\textit{prefer B}}. We instruct annotators vote \textbf{\textit{none}} if neither \textit{A} nor \textit{B} satisfies both \textit{minimal editing} and \textit{persona consistency}. See detailed guidelines in Appendix~B and supplementary materials.

Table~\ref{tab:human-results} shows that human annotators generally prefer GME to the baselines. The free-marginal $\kappa$ for each row is 0.66, 0.66, and 0.51 (\textit{substantial, substantial, and moderate agreement}).
The strongest baseline DeLorean-FT is only preferred in $8.4 \%$ cases. We observe that in most cases where DeLorean-FT wins, the original response is syntactically similar to the persona. 

\begin{table*}[t]
\centering
\small
\begin{tabular}{@{}lcccccccc@{}}
\toprule
~  & \multicolumn{4}{c}{Automatic evaluation} & \multicolumn{4}{c}{Human evaluation} \\
\cmidrule(lr){2-5} \cmidrule(l){6-9}
~  			                    & BLEU  & F1        & P-Score       & NLL \down & Knowledge & Empathy & Persona & Grammaticality \\
\midrule
TransferTransfo                 & 2.31  & 13.96     & \bf $+$ 67.4  & 4.54      & 0.0 \%  & 4.0 \% & \textbf{82.7} \% & 90.7 \% \\
\midrule
Blender-90M                     & 3.23  & 19.22     & $+$ 9.2       & 3.73      & 3.3 \%  & 29.3 \% & 20.7 \% & 96.3 \% \\
\quad $+$ edited by GME         & 3.10  & 19.02     & \bf $+$ 33.0  & 3.82      & 3.0 \%  & 29.0 \% & \textbf{66.3} \% & 88.3 \% \\
\midrule
Blender-90M w/o persona         & 2.98  & 18.91     & $+$ 0.8       & 3.63      & 8.3 \%  & 28.7 \% & 1.3 \% & 97.0 \% \\
\quad $+$ edited by GME         & 2.84  & 18.87     & \bf $+$ 29.4  & 3.78      & 7.7 \%  & 28.0 \% & \textbf{56.7} \% & 91.7 \% \\
\bottomrule
\end{tabular}
\caption{\label{tab:transfer-learning-bst} Automatic and human evaluation for the transferability to the test set of \textsc{BlendedSkillTalk}. NLL is computed using GPT-2. Free-marginal $\kappa$ for knowledge, empathy, persona, and grammaticality is 0.92, 0.70, 0.85, and 0.78 (\textit{almost perfect, substantial, almost perfect, and substantial agreement}).}
\end{table*}

\subsection{Behavioral Analysis}
\label{subsec:model-behavioral-analysis}
Using the metrics defined in Section~\ref{subsec:data-analysis}, we provide a behavioral analysis of the models. Results are shown in Table~\ref{tab:behavioral-analysis}. CycleGAN and UNMT have small \textit{add}, \textit{rm}, and $d(r^{e}, r^{o})$, which shows that they make little changes to the original response. For DeLorean-FT, larger mixture rates $\gamma_{\itm{mix}}$ have larger \textit{add}, \textit{rm}, and $d(r^{e}, r^{o})$, which is consistent with the observation in Section~\ref{subsec:grounded-editing-automatic-evaluation}. The large $d(r^{e}, r^{o})$ of DeLorean-FT also shows that this model behaves poorly at making minimal editing. GME has the most similar behavior with human references. Based on the observations in Section~\ref{subsec:grounded-editing-automatic-evaluation}-\ref{subsec:model-behavioral-analysis}, we conclude that GME is effective in making minimal edits that are targeted at persona-related expressions. By checking the outputs, we observe that GME and human references add persona information into the response in some cases, which may explain why GME and human references have positive $\Delta L$ (i.e., their predictions are longer than the original responses). 

\section{Transferability Experiment}
\label{sec:experiment-transferability}
We evaluate the transferability of GME to minimally edit existing responses on the test split of \textsc{BlendedSkillTalk} \cite{SmithWSWB20}. We also evaluate whether grounded minimal editing addresses the transferability challenges of grounded dialogue modeling (Q2 in Section~\ref{sec:introduction}). 

\subsection{Experimental Setup}
\label{subsec:transfer-setup}
In \textsc{BlendedSkillTalk}, each dialogue session is grounded on \textit{two persona sentences} and an optional \textit{knowledge topic}, and the distribution of responses is biased towards the mixture of \textit{displaying persona, using knowledge, and being empathetic}. Two types of existing responses are considered: 
\begin{itemize}
    \item Responses generated by a \textit{persona-agnostic} Blender-90M \cite{Roller20Blender}, which is trained on \textsc{BlendedSkillTalk} in which the persona sentences are removed. 
    \item Responses generated by the original \textit{persona-grounded} Blender-90M.
\end{itemize}
We compare the above two Blender-90M variants and GME-edited resposnes with TransferTransfo \cite{wolf19transfertransfo}, a pretrained dialogue model finetuned on \textsc{PersonaChat}. Note that GME is \textit{not} finetuned on \textsc{BlendedSkillTalk}. Also, conversations in \textsc{PersonaChat}, on which GME and TransferTransfo are trained, barely display knowledge and empathy. 

\subsection{Automatic Evaluation}
\label{subsec:transfer-automatic-evaluation}
We report BLEU and F1 \cite{MillerFBBFLPW17} computed with the human references. For persona consistency, we report the P-Score defined in Section~\ref{subsec:grounded-editing-automatic-evaluation}. To evaluate fluency, we report the word-level NLL evaluated by GPT-2 \cite{Radford2019GPT2}. 
The automatic evaluation uses the full 5482 test samples of \textsc{BlendedSkillTalk}. 

Table~\ref{tab:transfer-learning-bst} shows that P-Scores are largely improved after GME editing (from $9.2$ to $33.0$, and from $0.8$ to $29.4$). BLEU, F1, and NLL remain comparable to those before editing. Although TransferTransfo has the highest persona consistency, it has much poorer BLEU, F1, and NLL than GME. These results show that grounded minimal editing addresses the transferability issue faced by TransferTransfo. 

\subsection{Human Evaluation}
\label{subsec:transfer-human-evaluation}
We randomly sample 100 test samples for human evaluation. Three annotators evaluate if a response shows \textit{knowledge}, \textit{empathy}, \textit{persona consistency}, and \textit{grammaticality}. See detailed guidelines in in Appendix~B and supplementary materials.

Results are shown in Table~\ref{tab:transfer-learning-bst}. Free-marginal $\kappa$ for knowledge, empathy, persona, and grammaticality is 0.92, 0.70, 0.85, and 0.78 (\textit{almost perfect, substantial, almost perfect, and substantial agreement}), respectively. Results show that persona consistency is largely improved after GME editing, while the use of knowledge and empathy remain comparable to those before editing. TransferTransfo has the highest persona consistency, but it has much lower knowledge and empathy than the responses edited by GME. For example, only $4.0\%$ of TransferTransfo's responses show empathy, while the ratios are $29.0\%$ and $28.0\%$ for the GME-edited responses. We also notice a slight grammaticality drop after GME editing. However, the GME edited responses still achieve competitive or higher grammaticality scores comparing to Transfertransfo. In practice, the grammaticality scores can be easily improved using re-ranking approaches. In summary, GME largely improves the persona consistency of existing responses while preserving their use of knowledge and empathy, which addresses the transferability challenges faced by grounded dialogue models trained on \textsc{PersonaChat}, e.g., TransferTransfo. 

\section{Discussion}

As mentioned in Section~\ref{sec:related-work}, the term ``\textit{minimal}'' distinguishes our work from two-pass generation \cite{XiaTWLQYL17} and retrieval-augmented dialogue models \cite{WestonDM18,CaiWBTLLS19}. Generally, their objective can be formulated as $P(r|c, r')$ where $r'$ is a response either generated by the model itself or retrieved from the dataset. However, these works do not require $r$ and $r'$ to be a minimal editing pair. By contrast, we formulate $r^{e}$ and $r^{o}$ to be a minimal editing pair. To encourage \textit{minimal} editing, we construct response templates from the observed responses themselves, while these works derive templates from the $r'$ defined above. 

GME itself is also trained on a grounded dialogue dataset that has biased distribution. Thus, as we mentioned at the beginning of Section~\ref{sec:experiment-transferability}, we also need to evaluate the \textit{transferability} of GME. Section~\ref{sec:experiment-transferability} shows that GME editing only slightly changes the distribution of the responses generated by the Blender-90M variants, while the distribution of TransferTransfo's responses is further away from the human references. This observation suggests that \textit{minimally editing} out-of-domain responses is easier than \textit{generating} them. 

While we focus on the persona, other types of grounding, e.g., \textit{knowledge} and \textit{image}, remain to be explored. Many of GME's failure cases (see 
Appendix~E) contain grammatical errors or fail to correct contradictions, which could be addressed by improving the quality of response templates or incorporating stronger language model priors. 

\section{Conclusions}
\label{conclusions}
We propose a framework named \textit{grounded minimal editing} to address the \textit{transferability} challenges of grounded dialogue modeling, which include the distributional gap and the grounding gap. Our Grounded Minimal Editor (GME) model achieves minimal editing by disentangling and recombining persona-related and persona-irrelevant expressions. For evaluation, we present the \textsc{PersonaMinEdit} dataset with multiple human references. Experimental results show the effectiveness of GME for persona-grounded minimal editing. GME is also transferable to edit responses generated by pretrained dialogue models and improve their persona consistency while preserving their use of knowledge and empathy. 

\section*{Acknowledgements}
This work was supported by the National Science Foundation for Distinguished Young Scholars (with No. 62125604) and the NSFC projects (Key project with No. 61936010 and regular project with No. 61876096). This work was also supported by the Guoqiang Institute of Tsinghua University, with Grant No. 2019GQG1 and 2020GQG0005.

\bibliography{anthology,custom}
\bibliographystyle{acl_natbib}

\appendix
\section{Annotation Guideline (Simplified)}
\label{app:annotation-manual}
The following guideline is provided to AMT crowd-workers when collecting reference responses:

``We aim at building human-like agents that have their own personal background. We need your help to correct some responses that are irrelevant to or contradictory to the speaker's personal background. In each sample, you first see the dialogue history between the two speakers (Speaker1 and Speaker2), the original response by Speaker2, and the personal background of Speaker2. These background sentences are probably irrelevant to or contradictory to Speaker2's response. Your task is to minimally edit the response such that it shows the background. Two requirements should be satisfied: 1) The edited response should show the personal background. 2) By ``minimally edit'' we mean that the edited response should maintain the contents in the response that are not contradictory to the background sentences. We have pasted the original response into the answer blank, and please edit it directly.''

\section{Human Evaluation Guidelines}
\label{app:human-evaluation-manual}
We provide our human evaluation guidelines in \texttt{software.zip}, and we will make them public. We briefly summarize the guidelines here, and more details can be find in \texttt{software.zip}.

\subsection{Grounded Minimal Editing}
To make our task more comprehensible to human participants, we reformulate our task as a \textit{response correction} task. We define two types of mistakes made by a response: 1) \textit{contradict} and 2) \textit{ignore}. We first ask the participants \textbf{identify the type of mistake}, which encourages the participants to reason over persona-grounded dialogues. We specify two requirements to be satisfied by a \textit{good} correction: 1) the mistakes are corrected, and 2) minimal changes, i.e., all words that are not contradictory to the expected personal background should be maintained, and if more than four of them are not maintained, then this requirement is not satisfied. Given two corrections \textit{A} and \textit{B}, participants \textbf{vote \textit{prefer A}, \textit{none}, or \textit{prefer B}}. We ask them to choose \textbf{\textit{none}} if neither \textit{A} nor \textit{B} is a good correction.

\subsection{Transferability}
For each response, we instruct the participants to answer four questions:
\begin{itemize}
    \item \textbf{Knowledge (0 or 1)}. Does this response includes world knowledge? 0: no; 1: yes. World knowledge includes \textit{facts} and \textit{commonsense} (see \texttt{software.zip} for details).
    \item \textbf{Empathy (0 or 1)}. Do you think this response is showing empathy? 0: no; 1: yes. Showing empathy means being aware of or being sensitive to the \textit{feelings} or \textit{experience} of the person being talked to (see \texttt{software.zip} for details). 
    \item \textbf{Background occurrences}. There are two personal background sentences. How many of them are reflected by this response (examples omitted here)? Since we only care about if at least one of the personas are shown, \textbf{persona consistency} is \textbf{0} if the answer to this question is 0, and \textbf{1} if the answer is 1 or 2. 
    \item \textbf{Grammaticality (0 or 1)}. \ \ Is this response grammatical? 0: no; 1: yes. 
\end{itemize}

\section{Experimental Details}
\label{app:more-experimental-details}
Models are evaluated on the validation set for every 500 steps, based on the Average metric. The batch size is 32. We use Adam \cite{KingmaB14} with the initial learning rate $5\times 10^{-5}$ and gradient clip $1.0$. The learning rate decays by half when the Average metric does not improve for two validations, and training terminates after three decays. We detokenize the BPE tokens into English words for evaluation. More details for the Reproducibility Checklist are in \texttt{software.zip}. 

\section{Model and Baseline Details}
\label{app:model-details}
DeLorean-FT and our GME model use the GPT-2 \cite{Radford2019GPT2} as the backbone, initialized by Huggingface Transformers \cite{WolfDSCDMCRLFDS20} checkpoint \texttt{gpt2} (\texttt{DialoGPT-small} for DialoGPT). UNMT and CycleGAN are Transformers with DistilGPT-2 encoder and decoder, initialized with \texttt{distilgpt2}. The auxiliary response-to-persona module in CycleGAN is implemented as a two-layer Transformer, initialized by the first two layers of DistilGPT-2. 

\section{Data Samples and System Outputs}
\label{app:samples}
We provide several failure cases of GME in Table~\ref{tab:failure}. Table~\ref{tab:samples-1}-\ref{tab:samples-6} present some data samples and system outputs. 

\begin{table*}[t]
\centering
\small
\begin{tabular}{@{}lp{13cm}@{}}
\toprule
\multirow{5}{*}{Dialogue history}  		& Speaker 1:\quad hey , i love vegas , playing the slots ! \\
                                        & Speaker 2:\quad vegas is fun . i do not eat much so i can stay skinny and wear those jeans . \\
                                        & Speaker 1:\quad yeah , i like to grill outside in the summer ! \\
                                        & Speaker 2:\quad over my skinny jeans i like to wear leggings . summer grilling is the best . \\ 
                                        & Speaker 1:\quad i also like mowing my lawn when the suns out . \\
\midrule
Original response  			            & Speaker 2:\quad i put on the music when i mow the lawn or play video games . \\
\midrule
Editing persona  			        & i do not like music when i am working . \\
\midrule
Human reference 1                       & me too , but i do not like music when i mow the lawn or play video games . \\
                                        
Human reference 2                       & i do not like to put on the music when i mow the lawn or play video games . \\ 
Human reference 3                       & i like to mow the lawn or play video games when suns out , but i do not like music when i am working . \\
\midrule
GME (ours) // \textit{error}            & i put on the music when i mow the lawn or \red{play music} . i do not like music . // \textit{Contradiction} \\ 
\bottomrule
\toprule
\multirow{4}{*}{Context}  			    & Speaker 1:\quad hello . how are you today ?\\
                                        & Speaker 2:\quad good today . just cooking some mexican food . i cooking but am not very good .\\
                                        & Speaker 1:\quad oh i like mexican food , but my favorite food are cheeseburgers .\\
                                        & Speaker 2:\quad i inherited some money and bought a yacht to travel , i try different foods traveling . \\
\midrule
Original response  			            & Speaker 1:\quad i help out at a soup kitchen since i grew up poor . \\
\midrule
Editing persona  			        & i grew up with lots of cash . \\
\midrule
Human reference 1                       & i grew up rich but i help out at a soup kitchen . \\ 
Human reference 2                       & i have lots of cash , but i help out at a soup kitchen .  \\ 
Human reference 3                       & i help out at a soup kitchen since i grew up with lots of cash . \\
\midrule
GME (ours) // \textit{error}            & i help out at \red{a lot of cash} since i grew up . i grew up with lots of cash . // \textit{Grammatical error} \\
\bottomrule
\toprule
\multirow{9}{*}{Dialogue history}  		& Speaker 1:\quad hello , how are you today ? \\
                                        & Speaker 2:\quad hello , i am fine thanks and you ? \\
                                        & Speaker 1:\quad i am good because i love music and play it all the time . \\
                                        & Speaker 2:\quad ah that is nice ! i play softball in my free time . \\
                                        & Speaker 1:\quad nice , trabajo is my favorite spanish word . \\
                                        & Speaker 2:\quad i wish i had time to learn another language , but i am busy with work . \\
                                        & Speaker 1:\quad yeah i want to study french next . \\
                                        & Speaker 2:\quad since i have been fired from my last job i have been working in insurance . \\
                                        & Speaker 1:\quad that is pretty cool ! i love to study spanish . \\
\midrule
Original response  			            & Speaker 2:\quad i am a member of the army , served for 10 years now . \\
\midrule
Editing persona  			        & i am a school teacher , i teach middle school . \\
\midrule
Human reference 1                       & i am a school teacher and teach middle school , served for 10 years now . \\
                                        
Human reference 2                       & i am a school teacher for 10 years now and i teach middle school . \\ 
Human reference 3                       & i am a middle school teacher and i have teached for 10 years . \\
\midrule
GME (ours) // \textit{error}            & i am a teacher of the middle school . i teach middle school . // \textit{Repetition} \\
\bottomrule
\end{tabular}
\caption{\label{tab:failure} Failure cases}
\end{table*}

\begin{table*}[t]
\centering
\small
\begin{tabular}{@{}lp{13cm}@{}}
\toprule
\multirow{9}{*}{Dialogue history}  	    & Speaker 1:\quad hello , how are you doing ? \\ 
                                        & Speaker 2:\quad great , how are you ? i just finished watching one of my favorite documentaries . do you enjoy those ? \\
                                        & Speaker 1:\quad i am doing great , just tired . i just am unpacking boxes . i do not watch tv often .\\
                                        & Speaker 2:\quad did you just move ? i live here in pennsylvania with my husband .\\
                                        & Speaker 1:\quad yes , i bought my first house . i love pennsylvania , a lot of hills and very green .\\
                                        & Speaker 2:\quad good for you and congratulations on your new home ! \\
                                        & Speaker 1:\quad thank you ! so what do you do for work ?\\
                                        & Speaker 2:\quad i just started working as a personal assistant about three months ago . how about you ? \\
\midrule
Original response  			            & Speaker 1:\quad that sounds fun , i am a teacher at the public school . \\
\midrule
Editing persona  			        & i work at a place that cleans cars .\\
\midrule
Human reference 1                       & that sounds fun , i work at a place that cleans cars beside the public school . \\
Human reference 2                       & that sounds fun , i work at a place that cleans cars . \\
Human reference 3                       & that sounds fun , i am a teacher at the public school but i work as a car cleaner in part time . \\
\midrule
\midrule
UNMT ($0.1$)                            & that sounds fun , i am a teacher at the public school . \\
CycleGAN                                & that sounds fun , i am a teacher at the public school . \\
DeLorean-FT \ \ \\ 
\quad -- $\gamma_{\itm{mix}}=0.75$      & i work at a place that cleans cars . \\
\quad -- $\gamma_{\itm{mix}}=0.80$      & i work at a place that cleans cars . \\
\quad -- $\gamma_{\itm{mix}}=0.85$      & i work at a place that cleans cars . \\
\midrule
GME (ours)                              & that sounds fun , i am a car mechanic at the place . i work at a place . \\
\bottomrule
\end{tabular}
\caption{\label{tab:samples-1} Data sample and system outputs (\textit{correction}) }
\end{table*}

\begin{table*}[t]
\centering
\small
\begin{tabular}{@{}lp{13cm}@{}}
\toprule
\multirow{4}{*}{Dialogue history}  	    & Speaker 1:\quad hi . i do not like working as a car salesman . \\
                                        & Speaker 2:\quad i recently broke my arm so i am not working . \\
                                        & Speaker 1:\quad what happened ? it is hard to do anything with a broken arm .\\
                                        & Speaker 2:\quad i blame my skateboarding friends . \\
\midrule
Original response  			            & Speaker 1:\quad do you think 40 is too old to go back to school ? \\
\midrule
Editing persona  			        & i am seventy two years old . \\
\midrule
Human reference 1                       & do you think seventy two years old is too old to go back to school ?  \\ 
Human reference 2                       & do you think seventy two is too old to go back to school ? \\ 
Human reference 3                       & do you think seventy years old is too old to go back to school ? \\
\midrule
\midrule
UNMT ($0.1$)                            & do you think 40 is too old to go back to school ? \\ 
CycleGAN                                & do you think 40 is too old to go back to school ? i am seventy seventy twelve years old . \\ 
DeLorean-FT \ \ \\ 
\quad -- $\gamma_{\itm{mix}}=0.75$      & do you skateboard ? i am a seventy two year old . \\ 
\quad -- $\gamma_{\itm{mix}}=0.80$      & i am a seventy two year old man . \\ 
\quad -- $\gamma_{\itm{mix}}=0.85$      & i am a seventy two year old man . \\ 
\midrule
GME (ours)                              & do you think i am too old to go back to school ? i am seventy two years old . \\ 
\bottomrule
\end{tabular}
\caption{\label{tab:samples-2} Data sample and system outputs (\textit{correction}) }
\end{table*}

\begin{table*}[t]
\centering
\small
\begin{tabular}{@{}lp{13cm}@{}}
\toprule
\multirow{5}{*}{Dialogue history}       & Speaker 1:\quad hello hope your having a good night tonight ? i know i am . \\ 
                                        & Speaker 2:\quad that is good ! i am watching football highlights .\\
                                        & Speaker 1:\quad really what teams . and who are you rooting for .\\ 
                                        & Speaker 2:\quad i am a green bay packer fan . do you like football ? \\
                                        & Speaker 1:\quad no i am not a fan but my son is .\\
\midrule
Original response  			            & Speaker 2:\quad how many kids do you have ? \\
\midrule
Editing persona  			        & i don t like kids . \\
\midrule
Human reference 1                       & i don t like kids very much . how many kids do you have ? \\ 
Human reference 2                       & i don t have kids . how many kids do you have ? \\ 
Human reference 3                       & how many kids do you have ? i don t like kids . \\
\midrule
\midrule
UNMT ($0.1$)                            & how many kids do you have ? \\ 
CycleGAN                                & how many kids do you have ? \\ 
DeLorean-FT \ \ \\ 
\quad -- $\gamma_{\itm{mix}}=0.75$      & how many kids do you have ? \\ 
\quad -- $\gamma_{\itm{mix}}=0.80$      & i am a kid . i do not like kids . \\ 
\quad -- $\gamma_{\itm{mix}}=0.85$      & i am a kid . i do not like kids . \\ 
\midrule
GME (ours)                              & how many kids do you have ? i do not like kids . \\ 
\bottomrule
\end{tabular}
\caption{\label{tab:samples-5} Data sample and system outputs (\textit{add new information}) }
\end{table*}

\begin{table*}[t]
\centering
\small
\begin{tabular}{@{}lp{13cm}@{}}
\toprule
\multirow{2}{*}{Dialogue history}  	    & Speaker 1:\quad do you have any kids ? i have one in middle school .\\
                                        & Speaker 2:\quad i do not . my husband and i have chosen to not have any .\\
\midrule
Original response  			            & Speaker 1:\quad probably a smart decision , too many people on the planet . \\
\midrule
Editing persona  			        & i recently started to work online . \\
\midrule
Human reference 1                       & probably a smart decision , i recently started to work online because too many people on the planet .\\
                                        
Human reference 2                       & probably a smart decision , too many people on the planet that is why i recently started to work online . \\ 
Human reference 3                       & probably a smart decision , too many people on the planet . i recently started to work online . \\
\midrule
\midrule
UNMT ($0.1$)                            & probably a smart decision , too many people on the planet . \\ 
CycleGAN                                & probably a smart decision , too many people on the planet . \\ 
DeLorean-FT \ \ \\ 
\quad -- $\gamma_{\itm{mix}}=0.75$      & i am a computer science major . i am currently working online . \\ 
\quad -- $\gamma_{\itm{mix}}=0.80$      & i am a computer science major . i am currently working online . \\ 
\quad -- $\gamma_{\itm{mix}}=0.85$      & i am a computer science . i am currently working online . \\ 
\midrule
GME (ours)                              & probably a smart decision , too many people on the planet . i am working online now . \\ 
\bottomrule
\end{tabular}
\caption{\label{tab:samples-6} Data sample and system outputs (\textit{add new information}) }
\end{table*}


\end{document}